\documentclass[letterpaper]{article} 
\usepackage{aaai2026}  
\usepackage{times}  
\usepackage{helvet}  
\usepackage{courier}  
\usepackage[hyphens]{url}  
\usepackage{graphicx} 
\usepackage{natbib}  
\usepackage{caption} 
\usepackage{subcaption}
\usepackage{algorithm}
\usepackage{algorithmic}

\usepackage{newfloat}
\usepackage{listings}
\usepackage{times}
\usepackage{soul}
\usepackage{url}
\usepackage[utf8]{inputenc}
\usepackage{amsmath}
\usepackage{amsthm}
\usepackage{amssymb}
\usepackage{booktabs}
\usepackage{algorithm}
\usepackage{algorithmic}
\usepackage[switch]{lineno}

\usepackage{microtype}
\usepackage{booktabs} %
\usepackage{caption}
\usepackage{amsmath}
\usepackage{amssymb}
\usepackage{xcolor}
\usepackage{multirow}
\usepackage{tabularx}
\usepackage{tgcursor}
\usepackage{colortbl}
\usepackage{makecell}
\usepackage{graphicx}
\usepackage{subcaption}
\usepackage{natbib}
\usepackage{tikz}
\usepackage{amsmath}
\usetikzlibrary{arrows.meta, positioning}
\usepackage[most]{tcolorbox}
\usepackage{float}
\usepackage{xspace}
\tcbset{
  aibox/.style={
    width=\textwidth,
    top=10pt,
    colback=white,
    colframe=black,
    colbacktitle=black,
    enhanced,
    center,
    attach boxed title to top left={yshift=-0.1in,xshift=0.15in},
    boxed title style={boxrule=0pt,colframe=white,},
  }
}
\newtcolorbox{AIbox}[2][]{aibox,title=#2,#1}

\newtcolorbox[list inside=prompt,auto counter,number within=section]{prompt}[1][]{
    colbacktitle=black!60,
    coltitle=white,
    fontupper=\footnotesize,
    boxsep=5pt,
    enhanced,
    left=0pt,
    right=0pt,
    top=0pt,
    bottom=0pt,
    boxrule=1pt,
    breakable,
    #1
}

\DeclareCaptionStyle{ruled}{labelfont=normalfont,labelsep=colon,strut=off} 
\floatstyle{ruled}
\newfloat{listing}{tb}{lst}{}
\floatname{listing}{Listing}
\title{Tensorized Clustered LoRA Merging for Multi-Task Interference}

\author{
    Zhan Su\textsuperscript{\rm 1},
    Fengran Mo\textsuperscript{\rm 1}, 
    Guojun Liang\textsuperscript{\rm 2},
    JinghanZhang\textsuperscript{\rm 3}, 
    Bingbing Wen\textsuperscript{\rm 4},
    \\Prayag Tiwari\textsuperscript{\rm 2},
    Jian-Yun Nie\textsuperscript{\rm 1}\\
}

\affiliations{
    \textsuperscript{\rm 1}University of Montreal, Canada; 
   \textsuperscript{\rm 2}Halmstad University, Sweden; 
   \textsuperscript{\rm 3}Clemson University, USA;
   \textsuperscript{\rm 4}University of Washington, USA
   
    zhan.su@umontreal.ca
}

\usepackage{bibentry}

\begin{document}

\maketitle

\begin{abstract}
Despite the success of the monolithic dense paradigm of large language models (LLMs), the LoRA adapters offer an efficient solution by fine-tuning small task-specific modules and merging them with the base model. However, in multi-task settings, merging LoRA adapters trained on heterogeneous sources frequently causes \textit{task interference}, degrading downstream performance. To address this, we propose a tensorized clustered LoRA (TC-LoRA) library targeting to address the task interference at the \textit{text-level} and \textit{parameter-level}. At the \textit{text-level}, we cluster the training samples in the embedding space to capture input-format similarities, then train a specialized LoRA adapter for each cluster. At the \textit{parameter-level}, we introduce a joint Canonical Polyadic (CP) decomposition that disentangles task-specific and shared factors across LoRA adapters. This joint factorization preserves essential knowledge while reducing cross-task interference. 
Extensive experiments on out-of-domain zero-shot and skill-composition tasks-including reasoning, question answering, and coding. Compared to strong SVD-based baselines, TC-LoRA achieves +1.4\% accuracy on Phi-3 and +2.3\% on Mistral-7B (+2.3\%), demonstrating the effectiveness of TC-LoRA in LLM adaptation. 
\end{abstract}

\section{Introduction}

Parameter‑efficient fine‑tuning (PEFT) methods have emerged as practical alternatives to full fine‑tuning for adapting large language models (LLMs) to downstream tasks \citep{hu2021lora,houlsby2019parameter,lester2021power}. Among these, LoRA adapters \citep{hu2021lora} have gained particular popularity by learning low‑rank additive updates that efficiently capture task‑specific knowledge. Recent studies \citep{sheng2023s,ostapenko2024towards,huang2023lorahub} have shown that LoRA adapters can be fine‑tuned for diverse tasks—such as summarization, question answering, and classification—resulting in a LoRA library.

The growing availability of such pre‑trained adapters has motivated research on techniques to combine and reuse them effectively \citep{huang2023lorahub,stoica2024model}. Among these, model merging has emerged as a particularly appealing solution: it integrates multiple LoRA adapters into a unified multi‑task model without any additional training \citep{wortsman2022model,yadav2023ties,yang2024model,stoica2024model,gargiulo2025task}. This approach enables broad multi‑task capabilities in a training‑free manner, making it a compelling strategy for leveraging LoRA libraries at scale.

However, merging different task-specific LoRA adapters can introduce task interference, as the parameter updates learned for one task may conflict with those of another \citep{yadav2023ties}. At the parameter level, each LoRA adapter encodes task-specific transformations in its low-rank matrices. When these matrices are combined, their updates may point in different or even opposing directions in parameter space. To address this issue, many studies aim to reduce parameter redundancy, since overlapping or conflicting parameter updates are a primary cause of task interference in LoRA merging. For example, \texttt{Ties} \cite{yadav2023ties} reduces the parameter redundancy by selecting the top-$k$ most significant parameters. \texttt{DARE} \cite{yu2024language} resets the redundant parameters randomly to the pretrained values and rescales the remaining parameters by a factor proportional to the dropped ones to reduce interference among tasks. Most recently, researchers have explored singular value decomposition (SVD)-based merging of LoRA. The intuition is that SVD can decompose each LoRA adapter into orthogonal components and retain only the most significant singular vectors, which capture the core task-specific transformations \cite{lu2024twin,marczak2025no,gargiulo2025task}. For example, \texttt{TSV-merge} \cite{marczak2025no} applies the SVD decomposition to the LoRA matrix for each layer. This decomposition yields singular vectors and singular values that capture the most significant directions of variation within each layer. In this case, their approach can compress the LoRA to 10\%
of their original size while retaining 99\% of accuracy.

Despite the progress of LoRA merging methods in mitigating task interference, two key limitations remain. 1) Merging multiple LoRA adapters trained on different instruction datasets can lead to task interference, especially when the underlying tasks differ significantly in format, style, or input distribution. 2) While SVD-based decomposition can reduce parameter redundancy within each adapter, it processes adapters independently. As a result, redundant parameters may persist when two adapters share similar task-specific transformations.
To address these two issues, we propose a Tensorized clustered LoRA (TC-LoRA), which operates at both \textit{text-level} and the \textit{parameter level}. At the \textit{text-evel}, we observe that even tasks of different types can share underlying similarities. To exploit this, we cluster the instruction datasets based on their input embeddings, ensuring that semantically similar examples are grouped together. This clustering reduces the likelihood of task conflicts and promotes positive transfer across related tasks. At the \textit{parameter level}, we concatenate all LoRA adapters into a third-order tensor. Unlike recent SVD-based methods for mitigating task interference \citep{stoica2024model,gargiulo2025task}, which decompose each adapter independently, we adopt Canonical Polyadic (CP) decomposition. By jointly factorizing all task-specific matrices, CP decomposition captures shared and task-specific components in a unified manner. This joint decomposition helps disentangle task-specific factors from shared factors, reducing interference while preserving essential task information.

To rigorously evaluate LoRA merging methods, we focus on out-of-domain and skill composition tasks.
Out-of-domain evaluation is crucial because it tests whether the merged LoRA adapters can generalize beyond the specific domains they were trained on, which is essential for building adaptable LLMs.
Skill composition tasks, on the other hand, assess the model's ability to integrate and apply multiple learned capabilities in novel combinations — a key challenge in multi-task adaptation.
We conduct experiments on 10 zero-shot tasks covering these settings. The results show that LoRA merging substantially improves the backbone model’s average accuracy on downstream tasks. Compared to strong SVD-based LoRA merging baselines, our proposed TC-LoRA achieves a 1.4\% improvement on Phi-3 and a 2.3\% improvement on Mistral-7B.

In summary, our contributions are: 

\begin{itemize}

    \item We propose the TC-LoRA approach to reduce the task interference at the \textit{text-level} and \textit{parameter-level}. 
    
    \item The experimental results demonstrate that TC-LoRA can greatly improve the out-of-domain tasks, indicating that CP decomposition can jointly decompose the task matrix and help disentangle task-specific factors from shared factors, reducing task interference while preserving essential task information. 

    \item We will release the code and dataset and encourage further research in this direction.
\end{itemize}

\section{Preliminaries}

\subsection{CP Decomposition}
\begin{figure}[h]
    \centering
    \includegraphics[width=0.99\linewidth]{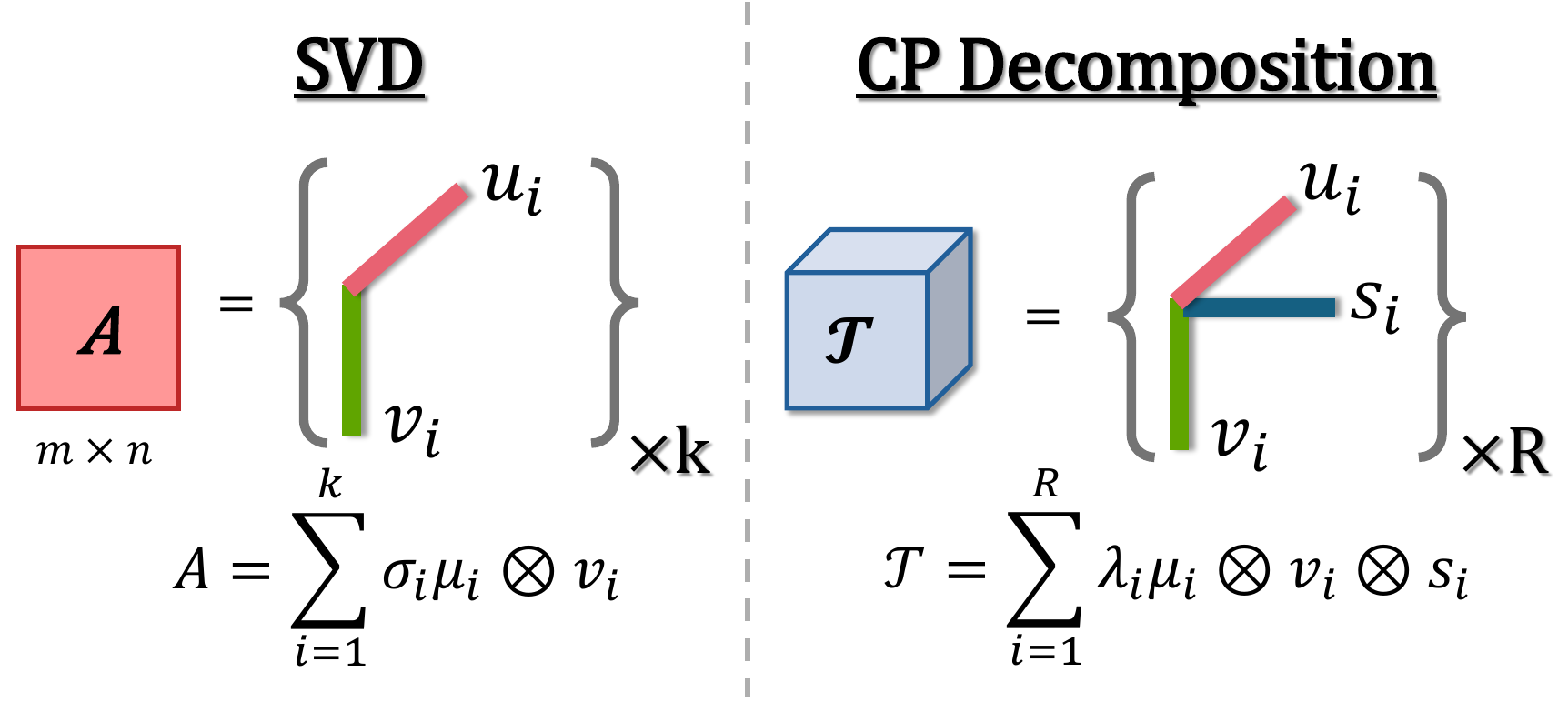}
    \caption{An illustration of SVD decomposition of matrix $A\in\mathbb{R}^{m\times n}$ and CP decomposition of third-order tensor $\mathcal{T}$.}
    \label{fig:svd_cp}
\end{figure}

Tensor decomposition aims to approximate a tensor through a set of low-rank factors with diverse applications. The most widely used decompositions are Tucker decomposition \citep{de2000multilinear} and CANDECOMP/PARAFAC (CP) decomposition \citep{tucker1966some}, both of which can be seen as a generalization of the matrix singular value decomposition (SVD) \citep{stewart1993early}. We compare the SVD and CP decomposition in Fig. \ref{fig:svd_cp}.   CP decomposition can be seen as a general SVD decomposition for higher-order tensors. In this way, a high-order tensor can be uniquely represented as the sum of a set of rank-one tensors. Given a third-order tensor $\mathcal{T}\in\mathbb{R}^{d\times r\times N}$, the CP decomposition is expressed as 
\begin{equation}
\label{eq:cp_decomposition}
    \mathcal{T}=\sum_{i=1}^R \lambda_i\mathbf{u}_i\otimes \mathbf{v}_i\otimes \mathbf{s}_i
\end{equation}

\begin{figure*}[htpb]
\small
    \centering
    \includegraphics[width=0.99\linewidth]{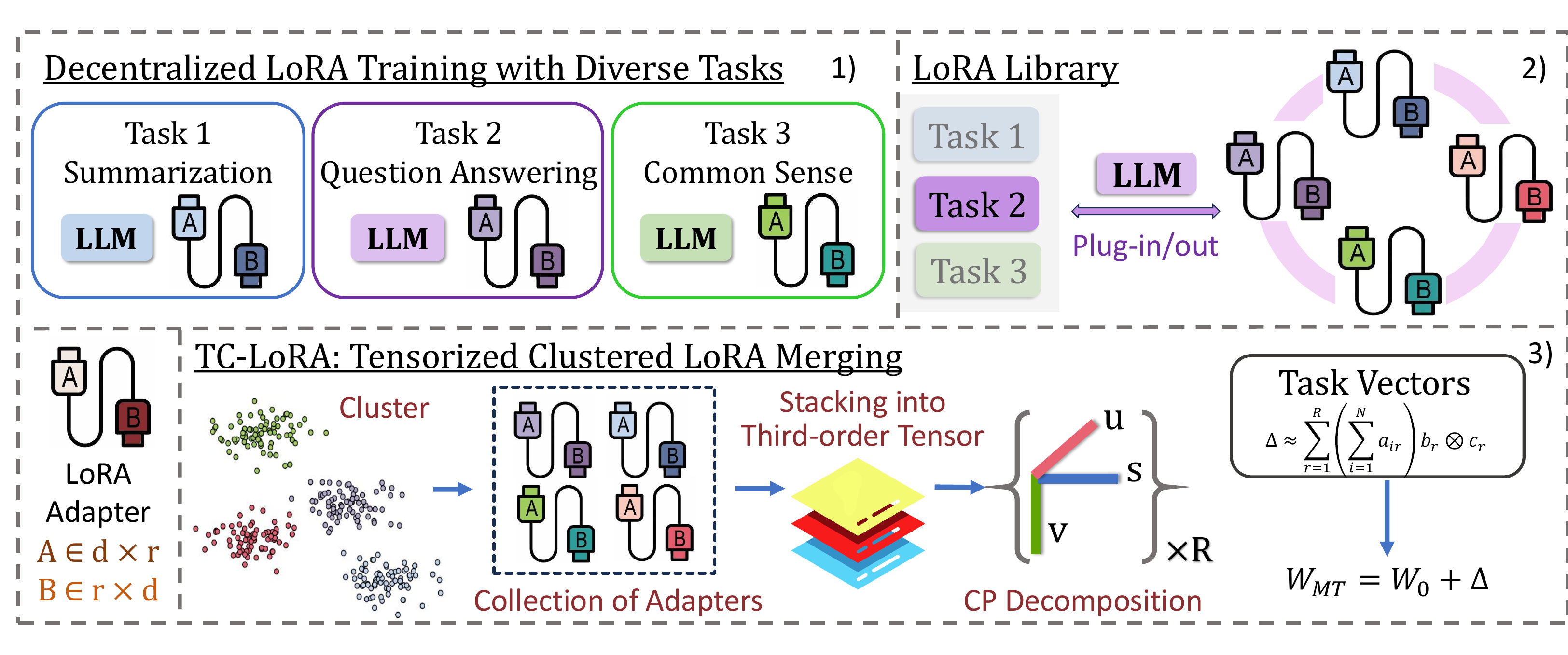}
    \caption{1) LoRA adapter from various tasks, each LoRA adapter is trained individually. 2) LoRA library aims to leverage the plug-in/out nature of LoRAs to offer the ability to add or remove knowledge from LLMs. 3) TC-LoRA. We first built the C-LoRA library. Then we stack the LoRA adapters into a third-order tensor to conduct the CP merging.}
    \label{fig:factorizing_lib}
\end{figure*}
where $R$ is the rank of CP decomposition. $\lambda_i$ is the scaling weight. $\mathbf{u}_i$, $\mathbf{v}_i$ and $\mathbf{s}_i$ are the factor vectors corresponding to the three modes of $\mathcal{T}$. The operator $\otimes$ denotes the outer product.

\section{Model}

In this section, we begin by introducing the zero-shot in a multi-task scenario. Following this, we introduce how the C-LoRA library is built.  Finally, we detail the integration of the C-LoRA to LLM with CP merging.

\subsection{Background}
\label{sec:problem}

Zero-shot generalization is a powerful concept in AI that enables models to adapt and perform new tasks by leveraging previously acquired knowledge. Given a model $f$ is trained on $n$ tasks $\{T_1,T_2,...,T_n\}$, the training process involves optimizing the parameters to minimize the combined loss over all these tasks. After training, the model $f$ is evaluated on new task $T_{\mathrm{new}}$ that it has never seen before. Let $D_i=\{(x_1,y_1),...,(x_n,y_n)\}$ represent the dataset for $T_i$, where $x_i$ are the inputs and $y_i$ are the corresponding outputs. For a new, unseen task $T_{\mathrm{new}}$ with dataset $\mathcal{D}_{\mathrm{new}}$, the model $f$ must predict outputs $\hat{y}$ for input $x$ without having been explicitly trained on $\mathcal{D}_{\mathrm{new}}$.

The success of zero-shot generalization heavily relies on the ability of a model to transfer knowledge from the training tasks to the new task. With the requirement to handle diverse tasks, the scalable solution LoRA library makes it straightforward to apply different adapters for different tasks without retraining the entire model. Formally, the weights $W_{\mathrm{MT}}$  of a multi-task model for $N$ tasks are obtained by aggregating the task-specific weight differences as follows:

\begin{equation}
    W_{\mathrm{MT}}^{(l)}=W_0^{(l)}+\alpha \frac{\Sigma_{i=1}^N\Delta_i^{(l)}}{N}
\end{equation}

where $W_0$ is the set of pretrained model weights. $\alpha$ is a scaling factor and $\Delta_i^{(l)}=A_i^{(l)}(B_i^{(l)})^T$ is the LoRA adapter for task $i$ at layer $l$. For brevity, we omit the layer index and refer to the matrix $\Delta_i^l$ at layer $l$ as $\Delta_i$. 

\subsection{Clusterized LoRA adapters(C-LoRA)}
\label{sec:factorizating}

To mitigate task conflicts at the \textit{text-level}, we train each \textit{expert} on a collection of similar tasks. This is achieved by clustering instructions and partitioning the training data into multiple clusters without requiring manual intervention. Clustering input texts into several groups can reduce task interference by allowing the model to adapt its task adapters or representations to distinct subsets of data with similar characteristics.  Formally, let $E(\cdot)$ denote a pretrained sentence encoder. For each data sample, the sentence representation of an instruction $I_i$ is computed as $e_i=E(I_i)$. We then apply the k-means algorithm to cluster all instruction embeddings $\{e_i\}$ in the training dataset into $K$ clusters. We train each LoRA adapter based on the instructions from each cluster. For each pre-trained weight $W_0$, we obtain corresponding LoRA adapters $\{A_iB_i,...,A_KB_K\}$.

\subsection{CP Merging}
\paragraph{Task Singular Vectors} 
\citep{gargiulo2025task} use SVD to decompose the layer task matrices and term the obtained singular vectors \textit{Task Singular Vectors (TSV)}, revealing low-rank properties and deeper insight into the inter-task interactions. Given two tasks $i$ and $j$, the tasks matrix $\Delta_i$ and $\Delta_j$ can written as:

\begin{align}
\Delta_i &= U_i \Sigma_i V_i^\top, \quad 
\Delta_j = U_j \Sigma_j V_j^\top
\end{align}

where $U_i$, $U_j$ are the left singular vectors and $V_i$, $V_j$ are the right singular vectors. $\Sigma_i$,$\Sigma_j$ are diagonal matrices of singular values. As shown in Fig. \ref{fig:svd_cp}, we can interpret SVD as a sum of rank-one matrices. For each task $i$, we get the best approximation of each task matrix $\Delta_i$ by retaining only the top-$k$ singular values and their corresponding vectors:

\begin{equation}
    \hat{\Delta}_i = \sum_{j=1}^{k} \sigma_j^i \, \mathbf{u}_j^i \otimes{\mathbf{v}_j^i}.
\end{equation}

where $\hat{\Delta}_i$ is the low-rank approximation of the task matrix for task $i$, obtained using the top-$k$ singular values $\sigma_{i,j}$, left singular vectors $\mathbf{u}_{i,j}$ and right singular vectors $\mathbf{v}_{i,j}$. SVD decomposes the task adapter separately, ignoring the interaction across different tasks. To address this, we propose a general task of singular vectors using Canonical Polyadic (CP) decomposition. 

\paragraph{Representing as a Third-Order Tensor}. We consider the task adapters $\Delta_i$ for all tasks $i$ as slices of a third-order tensor $\mathcal{T}$, where the dimensions could be interpreted as (tasks, input dimension, output dimension). Let use denote this tensor $\mathcal{T}\in\mathbb{R}^{d_{in}\times d_{out}\times N}$ where $N$ is the number of tasks, $d_{in}, d_{out}$ are the dimensions of each task matrix $\Delta_i$. Each $\Delta_i$ (a matrix of size $d_{in}\times d_{out}$) is a frontal slice of $\mathcal{T}$ along the task mode. So, we can concatenate the $\Delta_i$ matrices along the task dimension to form $\mathcal{T}$.

\paragraph{CP Decomposition Form}. In CP decomposition, the tensor $\mathcal{T}$ is approximated as a sum of rank-one tensors:

\begin{equation}
    \mathcal{T} \approx \sum_{r=1}^R \mathbf{a}_r \otimes \mathbf{b}_r \otimes \mathbf{c}_r
\end{equation}

where $R$ is the number of components (analogous to $k$ in SVD), $\mathbf{a}_r \in \mathbb{R}^N$ is a vector associated with the task mode, $\textbf{b}_r \in \mathbb{R}^d_{in}$ and $\textbf{c}_r \in \mathbb{R}^d_{out}$ are vectors associated with the row and column modes. For each task $i$, the $i$-th slice $\Delta_i$ of $\mathcal{T}$ would be:

\begin{equation}
\label{eq:cp_form}
    \Delta_i \approx \sum_{r=1}^R a_{ir} \mathbf{b}_r \otimes \mathbf{c}_r
\end{equation}

where $a_{ir}$ is the $i$-th element of $\mathbf{a}_r$, selecting the contribution of the $r$-th component for task $i$.

\paragraph{CP merging over the task mode}

To merge the $\Delta_i$ into a final $\Delta$, we sum over the task dimension (mode-1), the merged $\Delta$ would be:

\begin{equation}
    \Delta = \sum_{i=1}^N \Delta_i \approx \sum_{i=1}^N \sum_{r=1}^R a_{ir} \mathbf{b}_r\otimes \mathbf{c}_r
\end{equation}

Since the outer product $\textbf{b}_r \otimes \textbf{c}_r$ is the same for all $i$, and assuming $a_{ir}$ represents the weight for task $i$ in component $r$, we can factorize this as:
\begin{equation}
\label{eq:cp_merge}
    \Delta \approx \sum_{r=1}^R \left( \sum_{i=1}^N a_{ir} \right) \mathbf{b}_r \otimes \mathbf{c}_r
\end{equation}

where $\sum_{i=1}^N a_{ir}$ is the total contribution of the $r$-th component across all tasks. The final merged weights can be written as :

\begin{equation}
W_{\mathrm{MT}}=W_0+\sum_{r=1}^R \left( \sum_{i=1}^N a_{ir} \right) \textbf{b}_r \otimes \textbf{c}_r
\end{equation}

\begin{algorithm}[htpb]
\caption{Implementation of TC-LoRA building}
\algsetup{linenosize=\small}  
\begin{algorithmic}[1]  
    \STATE \textbf{Input: Multi-task $\{T_1,T_2,..., T_{\mathcal{T}}\}$ and data $\{\mathcal{D}_1,...,\mathcal{D_T}\}$} 
\STATE \textbf{Output: Merged weight $W_{\mathrm{MT}}$} 
\STATE $E=[]$
\STATE \textbf{Step 1: Obtain C-LoRA}
    \FOR{each sample $s$ in $\mathcal{D}_1,...,\mathcal{D_T}$ }
    \STATE embedding $e$ = SentenceEmbedding.encode($s$)
    \STATE $E$.append ($e$)
    \ENDFOR
\STATE Clusters $\{\mathcal{C}_1,...,\mathcal{C}_N\}$ = Kmeans(\textit{E})

    \FOR{$i = 1$ to $N$}
        \STATE  $A_k,B_k$=Train($\mathcal{C}_i$, LLM)
    \ENDFOR
   
\STATE \textbf{Step 2: CP Merging}
\STATE \textbf{Concatenate} the matrices:
\STATE $\mathcal{T}\leftarrow [\Delta_1 |\Delta_2|...|\Delta_N]$

\STATE  \textbf{Compute the CP Decomposition Form}

\STATE $\Delta_i \approx \sum_{r=1}^R a_{ir} \textbf{b}_r \otimes \textbf{c}_r$ \quad\quad (Eq.\ref{eq:cp_form})

\STATE \textbf{Reconstruct} the merged matrix:

\STATE  $\Delta \approx \sum_{r=1}^R \left( \sum_{i=1}^N a_{ir} \right) \textbf{b}_r \otimes \textbf{c}_r$ \quad\quad (Eq. \ref{eq:cp_merge})

\STATE \textbf{Construct} the merged model weights:
\STATE $W_{\mathrm{MT}}=W_0+\sum_{r=1}^R \left( \sum_{i=1}^N a_{ir} \right) \textbf{b}_r \otimes \textbf{c}_r$
\STATE \textbf{return} $W_{\mathrm{MT}}$
\end{algorithmic}
\label{al:factorizing}
\end{algorithm}

\section{Zero-shot Experiments}
\label{sec:experiment}
\begin{table*}[ht]
\small
\centering
\begin{tabular}{l|cccccccccc|c} 
\Xhline{1.6pt}
\multicolumn{1}{c|}{} & \multicolumn{3}{c}{\textbf{Common Sense}} & \multicolumn{4}{c}{\textbf{Question Answering}} & \multicolumn{2}{c}{\textbf{Coding}} & \multicolumn{1}{c}{\textbf{ Reasoning}} & \multicolumn{1}{c}{\textbf{AVG}} \\
\cmidrule(lr){2-4}  \cmidrule(lr){5-8} \cmidrule(lr){9-10}
\cmidrule(lr){11-11}
\multicolumn{1}{c|}{} & \textbf{Piqa} & \textbf{Wg} & \textbf{Hswag} & \textbf{Boolq} & \textbf{Obqa} & \textbf{ArcE} & \textbf{ArcC} & \textbf{HE} & \textbf{Mbpp} & \textbf{BBH} & \multicolumn{1}{c}{\textbf{AVG}} \\
\midrule

\rowcolor{lightgray}\multicolumn{12}{c}{\textbf{Phi-3 (3B) with LoRA library }} \\
\textbf{BASE} & 81.1 & 70.2 & 75.7 & 85.0 & 49.0 & 80.0 & 57.9 & 53.0 & 61.1 & 53.4 & \cellcolor[HTML]{FDE9E8}66.6\\ 

\textbf{Multi-task}   & 79.1&69.6&75.6&87.1&47.6&82.6&55.6&51.8&52.9& 52.1 &\cellcolor[HTML]{FDE9E8}65.4 \\

\textbf{Uniform} & 81.0&70.4&76.0&84.7&48.8&82.4&57.9&52.4&63.0&55.6&\cellcolor[HTML]{FDE9E8}67.2\\

\textbf{Ties-merging} & 80.8 & \textbf{70.9} & 75.4 & 80.8 & 46.0 & 85.8 & 60.8 & 54.9 & 61.5 & 51.9 & \cellcolor[HTML]{FDE9E8}66.9 \\
\textbf{Task Arithmetic} & 81.0 & 70.5 & 75.9 & 84.7 & 48.8 & 82.5 & 58.2 & 54.3 & 61.9 & 54.2 & \cellcolor[HTML]{FDE9E8}67.2\\
\textbf{Ties w/DARE} & 80.7 & 70.8 & 75.2 & 82.8 & 46.3 & 84.8 & 60.9 & 54.9 & 61.6 & 52.0 & \cellcolor[HTML]{FDE9E8}67.0  \\
\textbf{TA w/DARE} & 81.2 & 70.1 & 75.3 & 84.9 & 48.9 & 82.2 & 58.3 & 54.5 & 61.1 & 54.3 & \cellcolor[HTML]{FDE9E8}67.1 \\
\textbf{TSV Merging} & 81.1 & 70.8 & 75.8 & 86.0 & 49.0 & 84.3 & 60.4 & 55.0 & 61.2 & 54.0 & 67.7\cellcolor[HTML]{FDE9E8}\\
\midrule

\textbf{C-LoRA} & 81.5 & 69.5 & 76.0 & 86.6 & 48.8 & 86.7  & 61.8 & 57.3  & 65.3 & 55.9  & \cellcolor[HTML]{FDE9E8}68.9 \textcolor{red}{\scriptsize ↑1.2} \\
\textbf{TC-LoRA} & \textbf{81.7} & 70.0 & \textbf{76.0} & \textbf{86.6}  & 48.9 & \textbf{86.7} & \textbf{62.5}& \textbf{57.3}  & \textbf{65.4} & \textbf{56.0}  & \cellcolor[HTML]{FDE9E8}\textbf{69.1} \textcolor{red}{\scriptsize ↑1.4} \\

\midrule

\rowcolor{lightgray}\multicolumn{12}{c}{\textbf{Mistral (7B) with LoRA library}  } \\
\textbf{BASE}  & 81.1  & 66.5  & 78.8  & 82.2 & 44.6 & 68.9 & 49.6 & 28.0 & 47.5 & 47.9 & \cellcolor[HTML]{FDE9E8}59.5\\ 
\textbf{Multi-task}  & 81.9 & 68.6 & 79.5 & 84.6 & \textbf{50.4} & 84.8 & 60.0 & 24.4 & 47.5 & 49.2 & \cellcolor[HTML]{FDE9E8}63.1 \\

\textbf{Uniform}& 82.1 & 67.2 & 79.6 & 82.7 & 45.2 & 78.7 & 54.8 & 29.9 & 49.4 & 49.0 &  \cellcolor[HTML]{FDE9E8}61.9 \\

\textbf{Ties-merging} & 82.4 & 70.6 & 79.9 & 87.6 & 44.2 & 81.1 & 57.3 & \textbf{31.7} & 47.9 & 46.2 & \cellcolor[HTML]{FDE9E8}62.9 \\
\textbf{Task Arithmetic} & 82.3 & 67.1 & 79.5 & 82.4 & 45.3 & 78.4 & 54.8 & 30.0 & 49.4 & 49.1 &  \cellcolor[HTML]{FDE9E8}61.8\\
\textbf{Ties w/DARE} & 82.3 & 70.3 & 79.8 & 87.4 & 44.3 & 81.3 & 57.2 & 31.6 & 47.6 & 46.3 & \cellcolor[HTML]{FDE9E8}62.8\\
\textbf{TA w/DARE} & 82.2 & 67.3 & 79.5 & 82.2 & 45.4 & 78.3 & 54.9 & 30.1 & 49.5 & 49.2 & \cellcolor[HTML]{FDE9E8}61.9\\
\textbf{TSV merging} & 81.6 & 70.3 & 78.9 & 82.8 & 42.8 & 84.0 & 58.3 & 28.7 & 49.7 & 49.5 & 62.7\cellcolor[HTML]{FDE9E8}\\

\midrule

\textbf{C-LoRA}  & 82.4  & 71.4 & 81.2 & 87.6 & 49.2 & 86.0 & 60.5  & 29.3 & 50.0 & 50.0 & \cellcolor[HTML]{FDE9E8}64.9 \textcolor{red}{\scriptsize ↑2.2} \\ 
\textbf{TC-LoRA} & \textbf{82.6} & \textbf{71.7} & \textbf{81.6} & \textbf{87.8} & 49.3  & \textbf{86.0} & \textbf{60.6} & 29.3  & \textbf{50.2}  & \textbf{50.7} & \cellcolor[HTML]{FDE9E8}\textbf{65.0} \textcolor{red}{\scriptsize ↑2.3} \\ \hline

\Xhline{1.6pt}
\end{tabular}
\caption{10 downstream Zero-shot results based on Phi-3 (microsoft/Phi-3-mini-4k-instruct), and Mistral-7B (mistralai/Mistral-7B-v0.1) backbones.}
\label{tab:zeroshot_full}
\end{table*}

To test the effectiveness of our approach, we evaluate TC-LoRA on a variety of zero-shot task benchmarks. Specifically, we experiment with four broad classes of zero-shot tasks: (1) \textbf{Common Sense Reasoning}: WinoGrande \citep{sakaguchi2021winogrande}, HellaSwag \citep{zellers2019hellaswag}, and PIQA \citep{bisk2020piqa}. 2) \textbf{Question Answering}: BoolQ \cite{clark2019boolq}, OpenbookQA \cite{mihaylov2018can}, ARC-easy \cite{clark2018think}, and ARC-challenge \cite{clark2018think}. 3) \textbf{Coding}: HumanEval \cite{chen2021evaluating}, MBPP \cite{austin2021program}. 4) \textbf{General-Purpose Reasoning}: BBH \citep{suzgun2022challenging}.

\subsection{Comparison Methods}

We compare the following methods in the zero-shot benchmark. \textbf{BASE}: the base model without adaptation. \textbf{Multi-task}: A single expert LoRA finetuned on a joint training set \citep{prabhakar2024lora}.
\textbf{Uniform}: For each task, we train a LoRA adapter, then we merge the LoRA uniformly \citep{huang2023lorahub,chronopoulou2023adaptersoup}. \textbf{Task Arithmetic (TA)} \citep{ilharco2022editing}. TA subtracts the parameter values of the pre-trained model from those of the fine-tuned models, creating a set of "task-vectors." Then they linearly summed to create a merged model. \textbf{Ties-merging} \citep{yadav2023ties}: Ties improves TA by resolving the parameter interference between models when merging. It prunes low-magnitude weights and then only averages weights that share the dominant sign. \textbf{DARE} randomly drops fine-tuned weights and rescales the remaining ones to create sparse task-vectors. \textbf{TSV merging} use SVD decomposition to compress the LoRA and reduce the task interference based on the singular vectors.

\subsection{Experimental results}

\begin{figure}[t]
\small
\centering
    \includegraphics[width=0.40\textwidth]{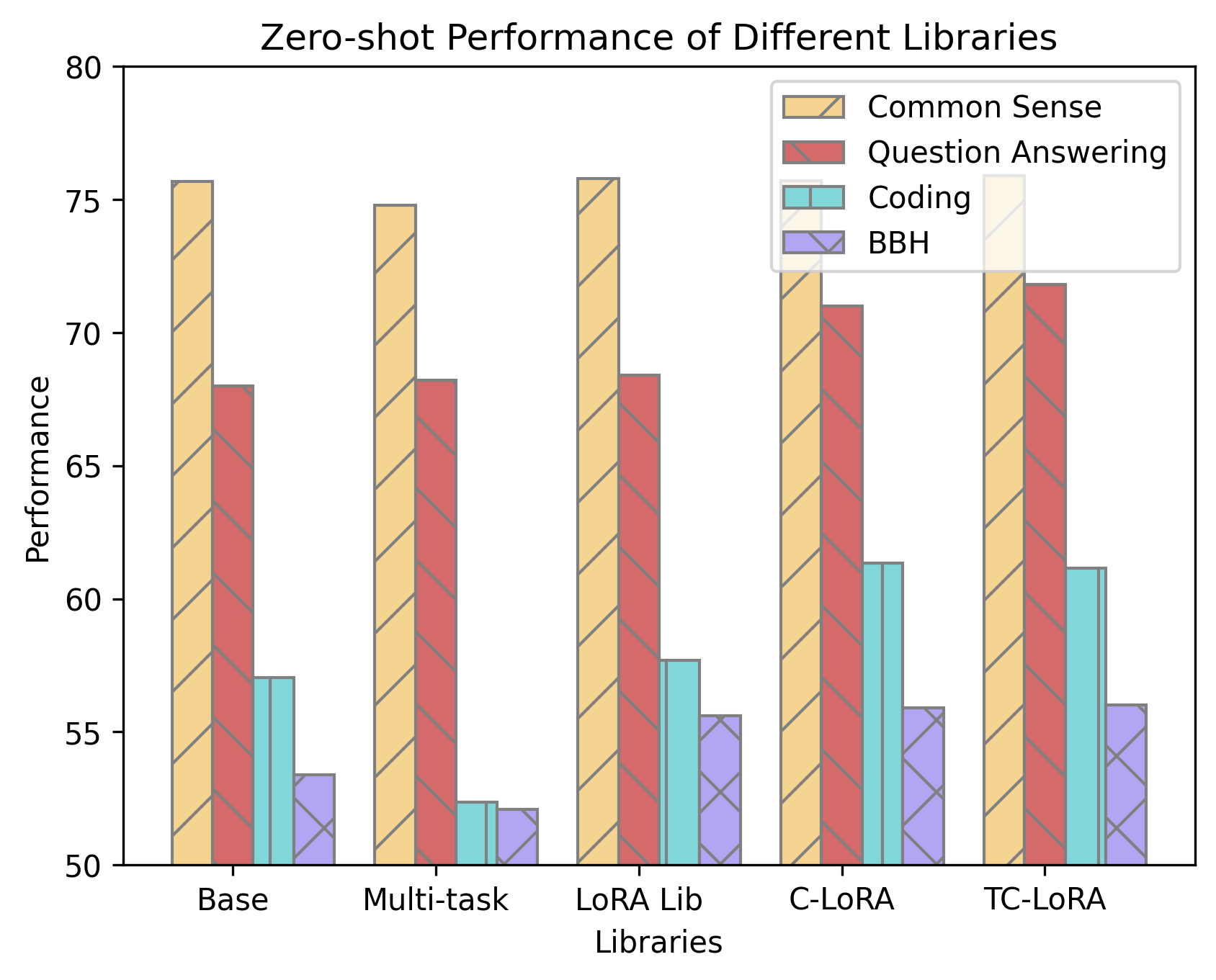}
\caption{Zero-shot results across different categories. The X-axis represents the various methods based on the LoRA adapter.}
\label{fig:performance_categories}
\end{figure}

Tab. \ref{tab:zeroshot_full} presents the average zero-shot accuracy results across 10 downstream tasks. For the Mistral (7B) model, the multi-task version achieves a notable improvement, raising the accuracy from 59.5\% to 63.1\%. However, the multi-task based on the Phi-3 model (65.4\%) underperforms relative to the base model (66.6\%), indicating that the multi-task training may encounter \textit{catastrophic forgetting} issue that newly learned parameters overwrite or interfere with knowledge acquired during prior training \citep{wang2024comprehensive}. LoRA library with uniform merging, consisting of 256 experts, which achieves accuracies of 67.2\% for Phi-3, and 63.1\% for Mistral (7B), representing improvements of 0.6\% and 2.6\% over the respective base models. Compared to the LoRA library, our clusterized approach C-LoRA, which utilizes only 10 experts, achieves accuracies of 68.9\% for Phi-3 and 64.9\% for Mistral-7B, outperforming the LoRA library with uniform by 1.7\% for Phi-3, and 3\% for Mistral-7B. This improvement underscores the effectiveness of clustering in enabling positive transfer during multi-task instruction tuning. Further analysis of the clustering method is provided in the discussion sections. Compared to C-LoRA, TC-LoRA approaches achieve a slight improvement over C-LoRA for Phi-3 and Mistral-7B. These results suggest that the redundancy persists within the LoRA library. The CP decomposition technique demonstrates the potential for mitigating this parameter redundancy.

We further analyze performance across different task categories (Fig. \ref{fig:performance_categories}). TC‑LoRA significantly improves over the LoRA library with uniform merging, with the largest gains observed in question answering and coding tasks. In contrast, improvements on common‑sense reasoning are minimal, suggesting that the impact of task interference varies by task type. In our setting, example‑level interference primarily affects question answering, while mitigating parameter‑level interference—compared to the C-LoRA baseline—yields notable gains in question answering performance. 

\section{Task Interference in Skill Composition Tasks}

\cite{prabhakar2024lora} have used LoRA merging to achieve the \textit{skill composition}. Skill here refers to specific capabilities that the LLM needs for customization to downstream use cases. For example, achieving a high score in GSM8k benchmark \cite{cobbe2021training} needs good \textit{commonsense} and \textit{arithmetic} skills.

\begin{figure}[h]
\centering
\begin{prompt}[title={GSM8k vs GSM8k-hard}] 
{
\textcolor{blue}{GSM8k}: A robe takes \textcolor{red}{2} bolts of blue fiber and half that much white fiber. How many bolts in total does it take?

Reference Answer: 18

\textcolor{blue}{GSM8k-hard}: A robe takes \textcolor{red}{2287720} bolts of blue fiber and half that much white fiber. How many bolts in total does it take?

Reference Answer: 3,431,580
}
\end{prompt}
\caption{The comparison between GSM8k and GSM8k-hard. The GSM8K uses a big number to replace the original number.}
\label{prompt-verifier}
\end{figure}

To explicitly examine how reducing parameter‑level task interference can improve performance, we evaluate our approach on hard math word problems. Compared to GSM‑8K \citep{cobbe2021training}, GSM8K‑hard \citep{gao2023pal} contains problems of similar types but with more complex arithmetic operations. \citet{gao2023pal} propose a program‑aided method in which an LLM first generates a program as an intermediate reasoning step and then executes it using a Python interpreter to obtain the final answer. This setting requires the LLM to be proficient in both mathematical reasoning (to analyze the problem) and coding (to translate the reasoning into executable code). Following \citet{prabhakar2024lora}, we train separate LoRA adapters for math and code skills. \textbf{The experimental setting is available in the supplementary materials}. 

\subsection{Results}

As depicted in Tab. \ref{tab:math-hard}, we evaluate the accuracy scores on the GSM-8k hard benchmark across various models. The base model achieves a modest accuracy of 0.043, highlighting its limitations in tackling GSM-hard tasks effectively. By training a LoRA adapter with the MathQA instruction dataset, we enhance the model's mathematical capabilities. Similarly, training with the Alpaca-code dataset boosts the model's coding proficiency. Combining both MathQA and Alpaca-code datasets in a single adapter yields an accuracy of 0.1349. Merging the math and coding skills through LoRA merging results in an accuracy of 0.1296, demonstrating the potential to integrate these abilities. Among the competing baselines, our proposed CP merging achieves the highest accuracy of 0.1569. Additionally, we assess the "invalid code" metrics across methods. Notably, the LoRA code approach reduces invalid code instances from 524 in the base model to 159, indicating improved code quality.

The results demonstrate that our CP merging effectively reduces task interference by integrating math and coding skills, as evidenced by the lower number of invalid codes (201) compared to the base model (524) and other methods.

\begin{table}[]
    \centering
    \begin{tabular}{l|cc}
    \toprule
    Model & ACC & Invalid code  \\
    \midrule
    \textbf{BASE}          & 0.0430 & 524  \\
    \textbf{LoRA(math)}   & 0.1175 & 562  \\
    \textbf{LoRA(code)}    & 0.0796 & 159 \\
    \textbf{Multi-task}    & 0.1349 & 413 \\
    \textbf{Uniform}       & 0.1296 & 217  \\
    \textbf{Ties-merging}  & 0.1060 & 223  \\
    \textbf{Task Arithmetic} & 0.1190 & 232 \\
    \textbf{Ties w/DARE}   & 0.1162 & 227 \\
    \textbf{TA w/DARE}     & 0.1200 & 247 \\
    \textbf{TSV Merging} & 0.1349 & 224 \\ 
    \midrule 
    \textbf{CP merging}       & \textbf{0.1569} & 201  \\
    \bottomrule
    \end{tabular}
    \caption{Evaluation results on MATH-hard tasks. "Invalid" is the percentage of invalid codes.}
    \label{tab:math-hard}
\end{table}

\section{Analysis}




\subsection{Rank $R$ of the CP decomposition.}
\label{sec:rank_analyse}

\begin{figure}[h]
    \centering
    \begin{subfigure}{0.23\textwidth}
        \centering
        \includegraphics[width=\textwidth]{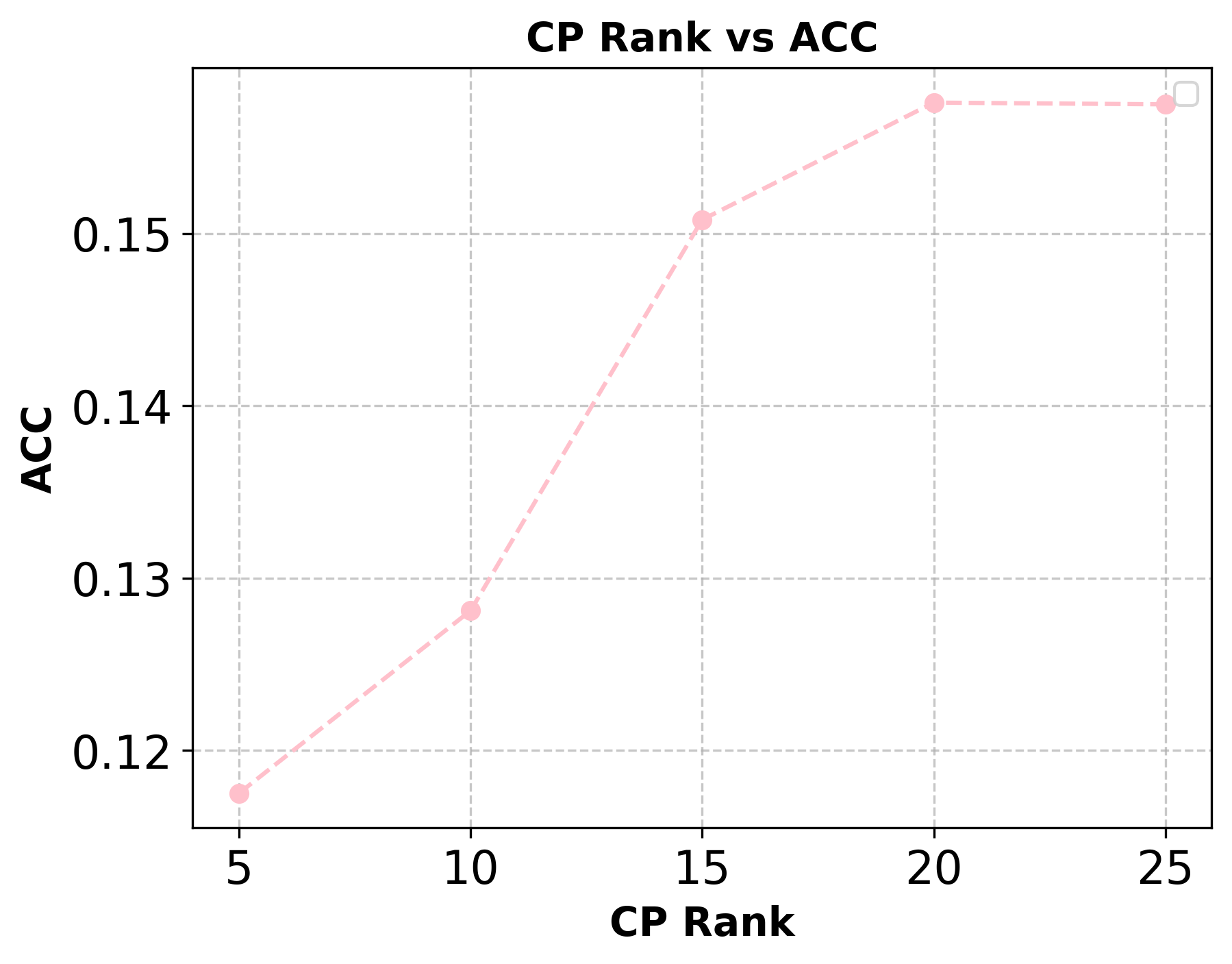}
       
    \end{subfigure}
    \hfill
    \begin{subfigure}{0.23\textwidth}
        \centering
        \includegraphics[width=\textwidth]{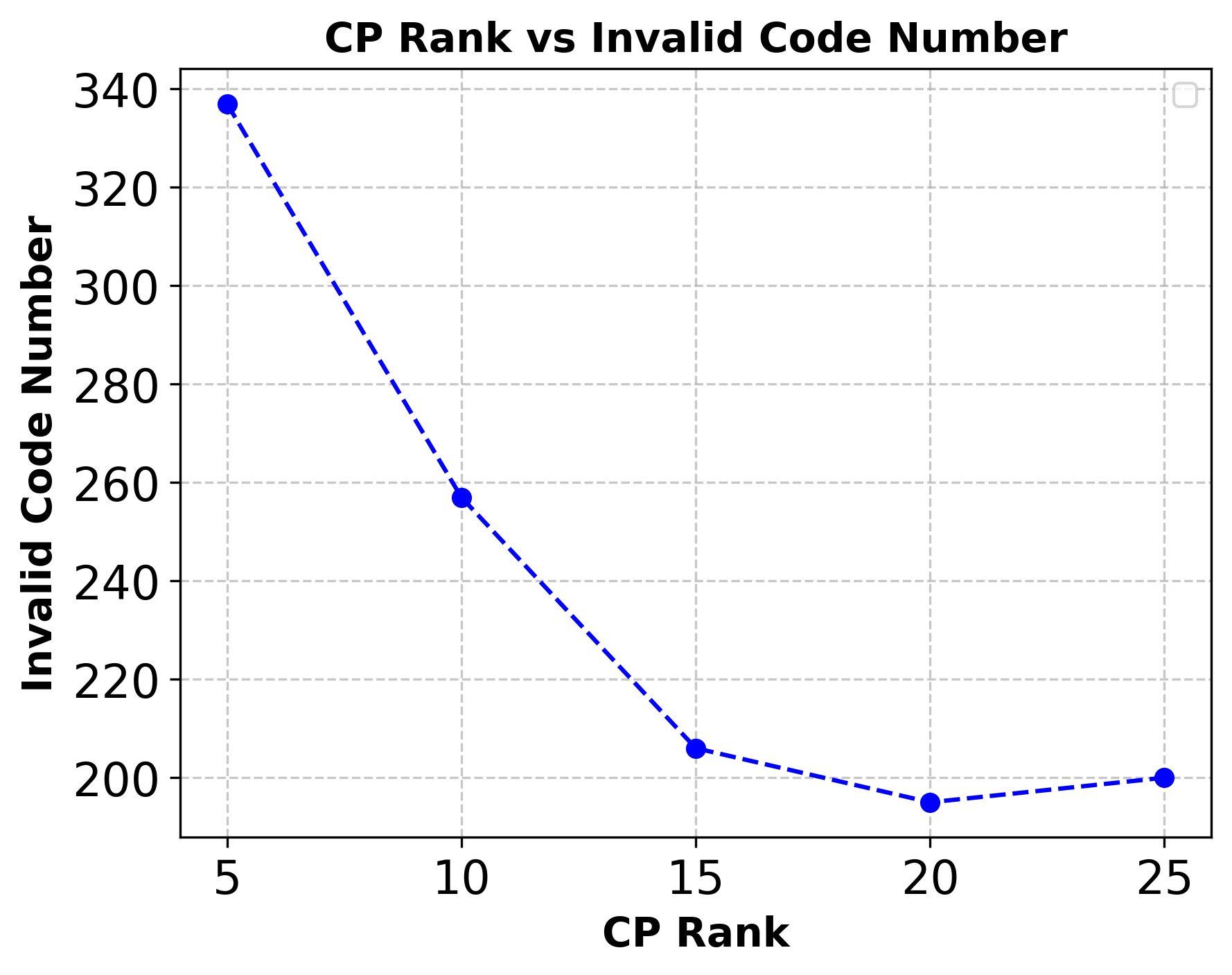}
    \end{subfigure}
    \caption{The impact of CP rank on model performance and generation quality for the GSM-8K benchmark.}
    \label{fig:CP_rank_performance}
\end{figure}

CP decomposition factorizes a tensor into a sum of $R$ rank‑one components. \citet{wang2023parameter} observed that increasing 
$R$ can improve performance on cross‑modal tasks. We investigate how the CP rank affects our method’s performance on the GSM8K‑hard dataset (Fig.\ref{fig:CP_rank_performance}). The left plot (“CP Rank vs. Accuracy”) shows accuracy rising from ~0.12 at rank 5 to a peak of ~0.15 at rank 20, with no further gains beyond rank 25. This suggests that higher ranks capture richer patterns, but benefits plateau after rank 20. The right plot (“CP Rank vs. Invalid Code”) shows invalid code outputs dropping from ~340 at rank 5 to ~200 at rank 20, with little improvement at higher ranks. Overall, increasing the CP rank improves both accuracy and output validity up to an optimal point (around rank 20), after which returns diminish—highlighting the importance of selecting an appropriate rank to balance performance and efficiency in mitigating task interference.

\subsection{Clustering to reduce the task interference in text-level}
\label{sec:clustering}

\begin{figure}[h]
    \centering
    \begin{subfigure}{0.23\textwidth}
        \centering
        \includegraphics[width=\textwidth]{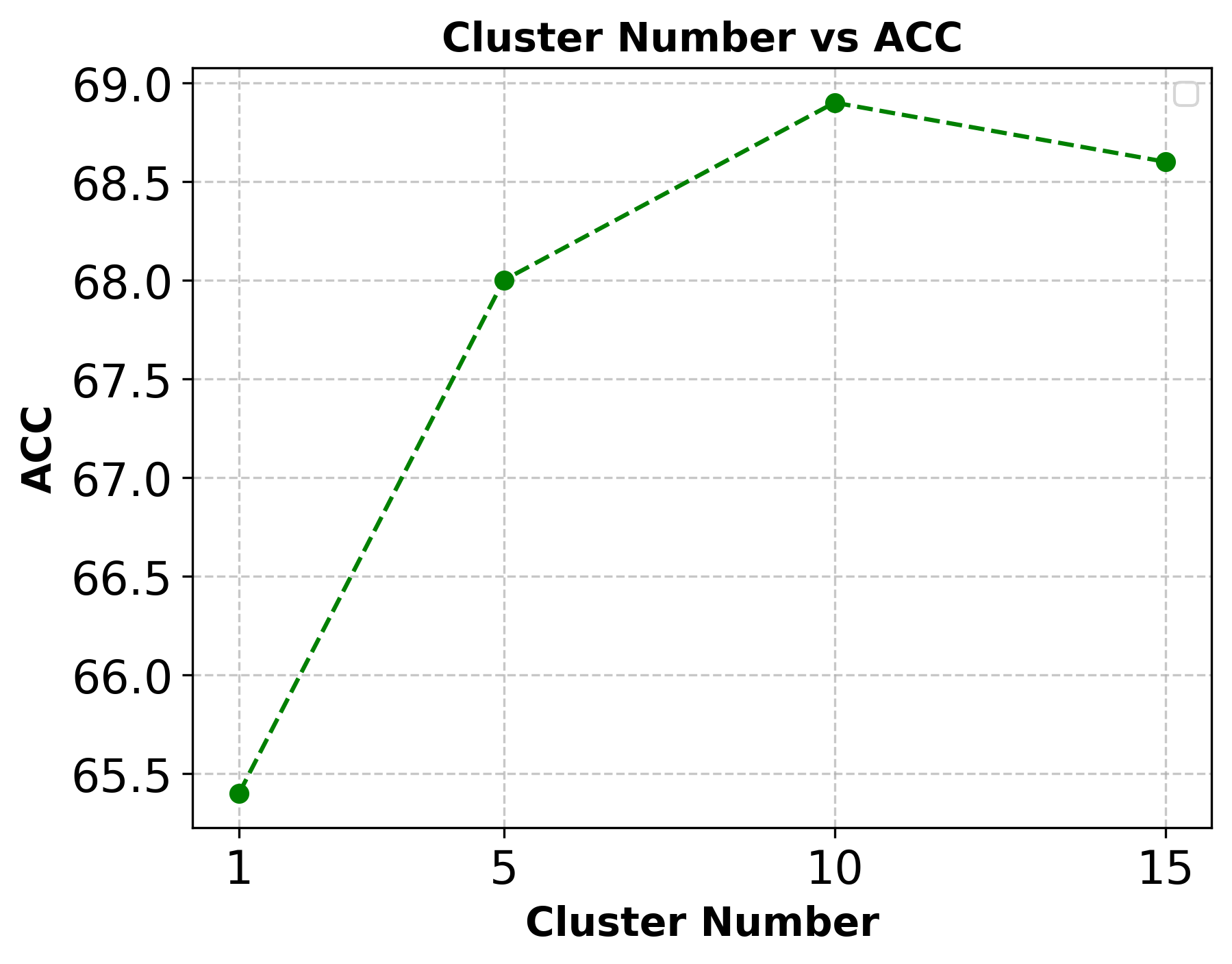}
       
    \end{subfigure}
    \hfill
    \begin{subfigure}{0.23\textwidth}
        \centering
        \includegraphics[width=\textwidth]{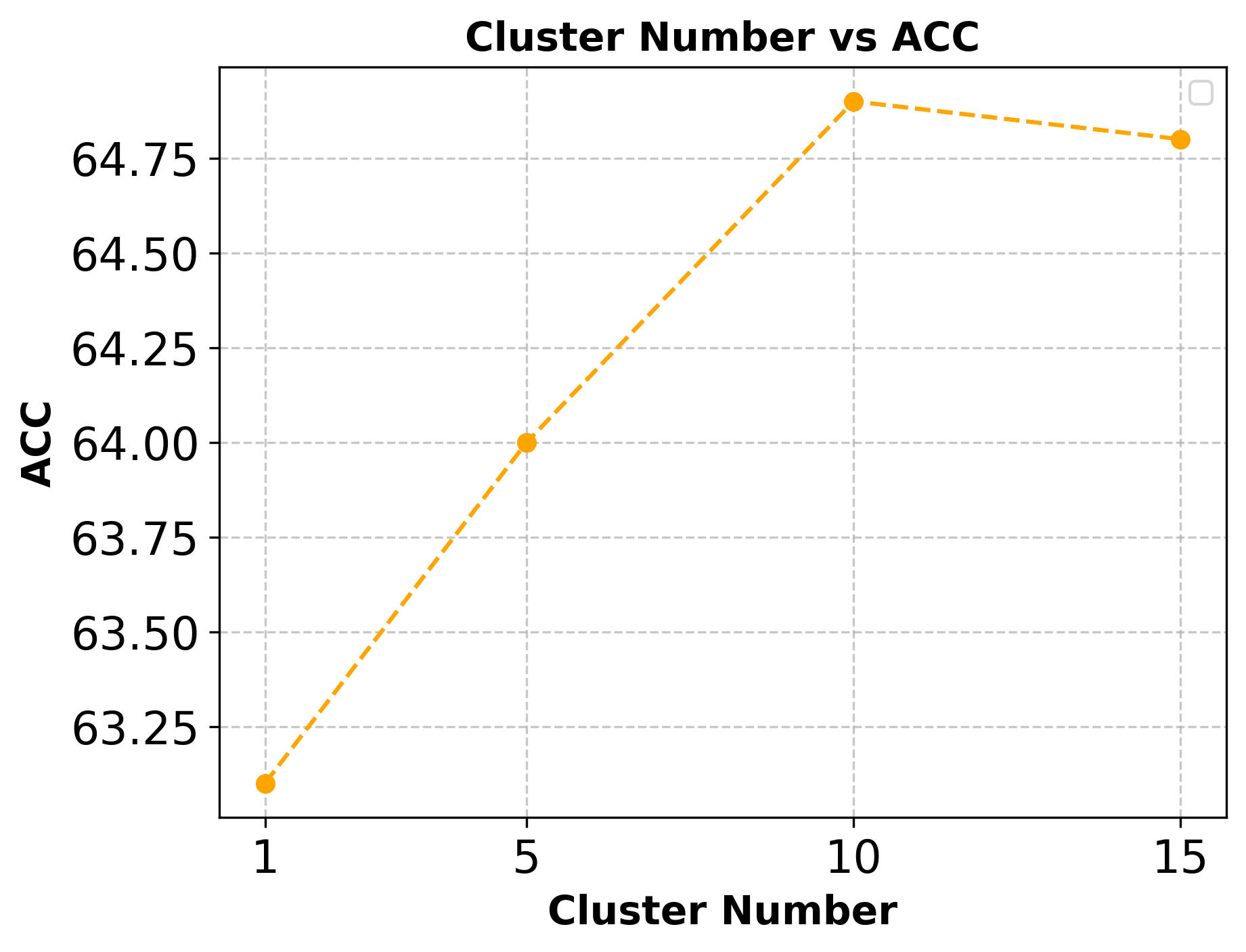}
    \end{subfigure}
    \caption{The number of clusters and accuracy (ACC) using Phi-3 3B (left) and Mistral 7B (right).}
    \label{fig:cluster_performance}
\end{figure}

The Fig. \ref{fig:cluster_performance} shows the relationship between the number of clusters and accuracy (ACC) using two different model backbones: Phi-3 3B (left) and Mistral 7B (right). For the Phi-3 3B backbone, accuracy increases from approximately 65.4\% at 1 cluster to a peak of 68.9\% at 10 clusters, stabilizing thereafter up to 15 clusters. In contrast, the Mistral 7B backbone shows a more gradual improvement, rising from 63.1\% at 1 cluster to 64.9\% at 10 clusters. We think it is still a challenge to determine the ideal number of clusters. Further investigation with more sophisticated clustering techniques could provide deeper insights into the relationship between performance and the number of clusters.




\subsection{CP Merging for Task Interference in parameter-level} 
\label{sec:cp_merging}

To study the task interference at \textit{parameter-level},  \cite{gargiulo2025task} introduces a score of task interference (termed as \textit{Singular Task Interference}) based on the interplay of TSVs from different tasks:

\begin{equation}
    \textbf{STI}\left( \left\{ \Delta_i \right\}_{i=1}^{N} \right) 
= \left\| \left( U^\top U - I \right) \Sigma \left( V^\top V - I \right) \right\|_1
\end{equation}

They assume that higher inner product values for $U^TU$ and $V^TV$ imply a higher likelihood \cite{gargiulo2025task}. The intuition is that overlapping singular vectors suggest shared features in the weight space across tasks. Inspired by this, we can reformulate the STI with CP decomposition. To define a CP-based STI metric, we need to assess the interference caused by the interplay of these factor matrices across tasks. The intuition remains that overlapping components (shared features in the weight space) lead to interference when merged. We can adapt the STI by considering the alignment and orthogonality of the factor matrices $a_r$, $b_r$, and $c_r$ across tasks. Then we define the CP-based Singular Task Interference (CP-STI) as:

\begin{equation}
\label{eq:cp_sti}
  \mathbf{CP-STI}(\{\Delta_i\}_{i=1}^N) = \|(A^T A - I) \circ (B^T B - I) \circ (C^T C - I)\|_1
\end{equation}

where $A=[a_1, a_2, \ldots, a_R] \in \mathbb{R}^{N \times R}$ is the matrix formed by stacking the task-mode factor vectors $a_r$, $B=[b_1, b_2, \ldots, b_R] \in \mathbb{R}^{d \times R}$ is the matrix of row-mode factors, $C= [c_1, c_2, \ldots, c_R] \in \mathbb{R}^{d \times R}$ is the matrix of column-mode factors, $\circ$ denotes the Hadamard (element-wise) product, $I$ is the identity matrix of appropriate dimension (e.g., $R \times R$). The CP-STI metric captures interference by evaluating how the factor matrices $A$, $B$, and $C$ deviate from being mutually orthogonal. Overlapping factors (high inner products) suggest shared features across tasks, which can introduce interference when the decomposed representations are merged, potentially degrading individual task performance.

\begin{figure}[t]
\centering
\includegraphics[width=0.45\textwidth]{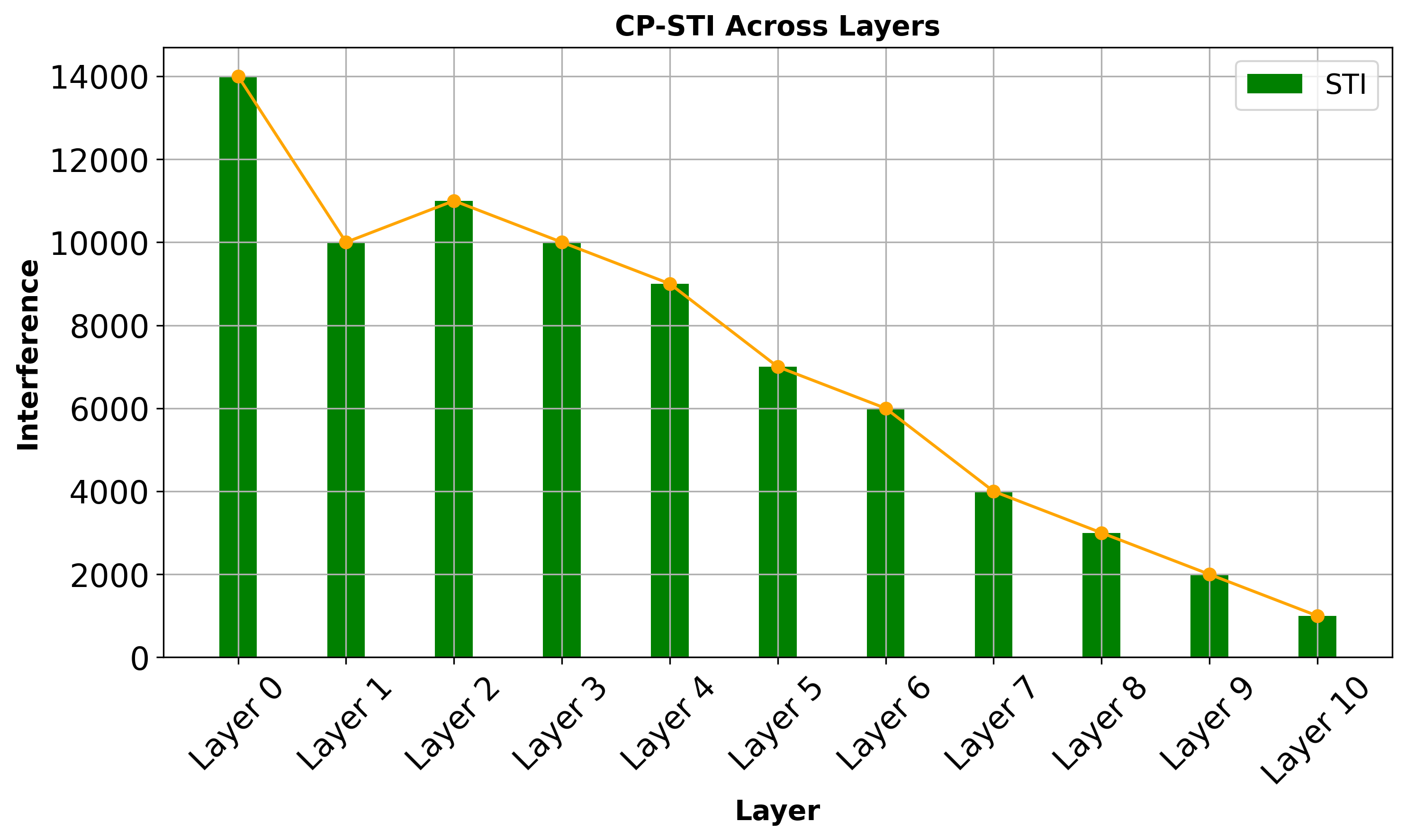}
\caption{CP-STI across 10 layers in the Mistral for Math and Code LoRA.}
\label{fig:cp_str_interference}
\end{figure}

We study the task interference on each layer according to Eq. (\ref{eq:cp_sti}). As shown in the Fig. \ref{fig:cp_str_interference}, the interference is higher in the lower layer and decreases in the deeper layers. This may indicate that the lower layers capture common features and deeper layers are more specialized for specific tasks.

\section{Related Works}

\noindent\textbf{PEFT}.  Parameter-efficient fine-tuning (PEFT) methods facilitate efficient adaptation of LLMs without updating all the training parameters, thereby reducing the memory and computation cost \citep{adapter_layer,zhang2023llama,zaken2021bitfit,guo2020parameter,li2021prefix,lester2021power}. \citet{hu2021lora} proposes a method named LoRA that parameterizes incremental weights $\Delta$ as a low-rank matrix by the product of the down projector matrix and up projector matrix. \citet{zhang2023adaptive} proposes Adaptive Low-Rank Adaptation (AdaLoRA), a new method that dynamically allocates the parameter budget among weight matrices during LoRA-like fine-tuning \citep{zhang2023adaptive}. \citet{liu2024dora} introduces a new approach DoRA to investigate the inherent differences between full fine-tuning and LoRA. 

\noindent\textbf{LoRA library}.
AdapterSoup \citet{chronopoulou2023adaptersoup} trains each adapter for each domain and performs weight-space averaging of adapters trained on different domains. \cite{huang2023lorahub} introduces LoRAhub to aggregate the LoRA modules trained on diverse tasks. AdapterFusion \citep{pfeiffer2020adapterfusion} proposes a two-stage algorithm that leverages knowledge from multiple tasks. Similarly to LoRAhub, a group of task-specific adapters learn to encapsulate the task-specific information, and in the second stage, a fusion layer combines the trained adapters. \citet{ostapenko2024towards} propose a model-based clustering library building methods and an Arrow routing function to reuse the LoRA library. \citet{zhao2024retrieval} propose a retrieval-augmented mixture of LoRA
Experts (RAMoLE) that 
adaptively retrieves and composes multiple LoRAs according to the input prompts. 

However, the previous LoRA library ignores the task interference from the \textit{text-level}. We study this using embedding clustering in our paper. 

\noindent\textbf{Model Merging}.
Model merging integrates the  weights of multiple task-specific models into a single multi-task
model \cite{davari2024model,ilharco2022editing,matena2022merging,wortsman2022model,yadav2023ties,yu2024language}. \cite{ilharco2022editing} presents the \texttt{task vectors}, which are the weight differences between pretrained models and the finetuned models. We can merge the task vectors to obtain a merged multi-task model. \texttt{Ties} \cite{yadav2023ties} reduces the parameter redundancy by selecting the top-$k$ most significant parameters and then constructing a sign vector based on the majority sign. \texttt{DARE} \cite{yu2024language} resets the redundant parameters randomly to the pretrained values and rescales the remaining parameters by a factor proportional to the dropped ones to reduce interference among tasks. \texttt{Fisher Merging} \cite{matena2022merging} and \texttt{RegMean} \cite{reg_mean} merge models by performing weighted averaging, utilizing the Fisher information matrix and inner product of input vectors. \texttt{Model Breadcrumbs} focuses on merging only significant weights by discarding outliers and both minor and large perturbations in the fine-tuned parameters \cite{davari2024model}. \texttt{TwinMerging} \cite{lu2024twin} propose Twin-merging to address the task interference. They use SVD to reduce the redundancy and dynamically merge shared and task-specific knowledge based on the input. 

Unlike the existing approaches, our method focuses on reducing the task interference for LoRA adapters from \textit{text-level} and \textit{parameter-level}. 

\noindent\textbf{Merging using SVD}.
\texttt{TSV} \cite{gargiulo2025task} shows that SVD decomposition can reduce the parameter redundancy, and merging the singular values can compress the parameters and reduce the task interference. \texttt{ISO-C} \cite{marczak2025no} propose an isotropic merging framework that flattens the isotropic merging framework to flatten singular value spectrum of the task matrix, enhancing alignment and reducing task interference. \texttt{Knots} \citep{stoica2024model} uses the SVD to jointly transform the weights of different LoRA models into an aligned space. 

Instead of decomposing the task matrix separately, we concatenate the task matrix into a third-order tensor and then decompose it using CP decomposition.

\section{Conclusion}

In this paper, we introduced TC‑LoRA, a framework for mitigating task interference in LoRA merging at both the text level and parameter level. At the text level, TC‑LoRA clusters training instances by input‑format similarity and trains specialized LoRA adapters for each cluster, thereby reducing example‑level interference. At the parameter level, we proposed a joint Canonical Polyadic (CP) decomposition that factorizes multiple LoRA adapters simultaneously, disentangling task‑specific and shared components. This joint decomposition preserves essential task knowledge while reducing cross‑task interference—an advantage over conventional SVD‑based merging methods. The extensive experimental results indicate that TC-LoRA can obtain the best results compared to other strong baselines on Phi-3 and Mistral. This demonstrates the potential of our LoRA library approach to enable more efficient and scalable implementations of post-training for LLMs.


\appendix

\section{Preliminaries}

\subsection{Tensor}
A tensor $\mathbf{A}$ is a multidimensional array of elements (called \emph{components}) of $\mathbb{R}$, each denoted by its integer coordinates in the array; e.g., for a two-dimensional array, the component at position $i,j \in \mathbb{N}$ is denoted $A_{ij}$.
 The \emph{order} of a tensor is how many indices it has (e.g., a vector $v$ is a first-order tensor, a matrix $M$ is a second-order tensor, etc.). The \emph{dimension} of a tensor refers to the number of values that a particular index (or so-called \textit{mode}) can take, e.g., the dimension of $B \in \mathbb{R}^{I_1 \times I_2 \times I_3}$ is $I_1 \times I_2 \times I_3$.

\subsection{LoRA adapters}
LoRA is a recently proposed adapter architecture that achieves a competitive balance between performance and parameter efficiency \citep{hu2021lora}. In each layer, for each linear transformation corresponding to the query ($q$), key ($k$), value ($v$), and output ($o$) of the self-attention layers, LoRA modifies the base model parameters as follows:
\begin{equation}
h = W_0 x + s\cdot A (B)^\top x
\label{eqn:lora}
\end{equation}

where $W_0$ are the (frozen) weights of the pre-trained model. $A, B \in \mathbb{R}^{d \times r}$ are low-rank learnable parameters and $s\ge 1$ is a tunable scalar hyperparameter.

\section{Experimental Setting}
\label{sec:exp_setting}

To enhance the LLM with various abilities for the above tasks, we construct a collection of LoRA adapters on diverse tasks. We use the FlanV2, an instruction-tuning dataset designed to scale both task diversity and model size, which has been shown to improve model performance significantly \citep{chung2024scaling}. We train each LoRA adapter independently, enabling it to specialize in a specific task. Finally, we select 256 tasks \citep{longpre2023flan}, with the majority derived from P3 \citep{bach2022promptsource} and Flan2021 \citep{wei2021finetuned}. 

To evaluate the effectiveness of our approach across different backbone models, we conduct experiments using Phi-3 \citep{abdin2024phi}, and Mistral-7B(mistralai/Mistral-7B-v0.1) \citep{jiang2023mistral}. In all cases, we apply LoRA adapters exclusively to the attention layers. Unless stated otherwise, our multi-task training and single-task adaptation scenarios employ a LoRA rank of 4, a dropout rate of 0.05, and a learning rate of 1e-4. Each LoRA adapter is trained for 5 epochs on either an H100 or A100 GPU. For the clustering step, we use the \texttt{sentence-transformers/sentence-t5-xxl} model for encoding. Given the typically large size of the dataset, we sample 20\% of the data to train the k-means algorithm and then use the resulting clusters to predict labels for all samples.

\section{Cluster based on Input text or LoRA parameters}

\begin{figure}[h]
    \centering
    \includegraphics[width=\linewidth]{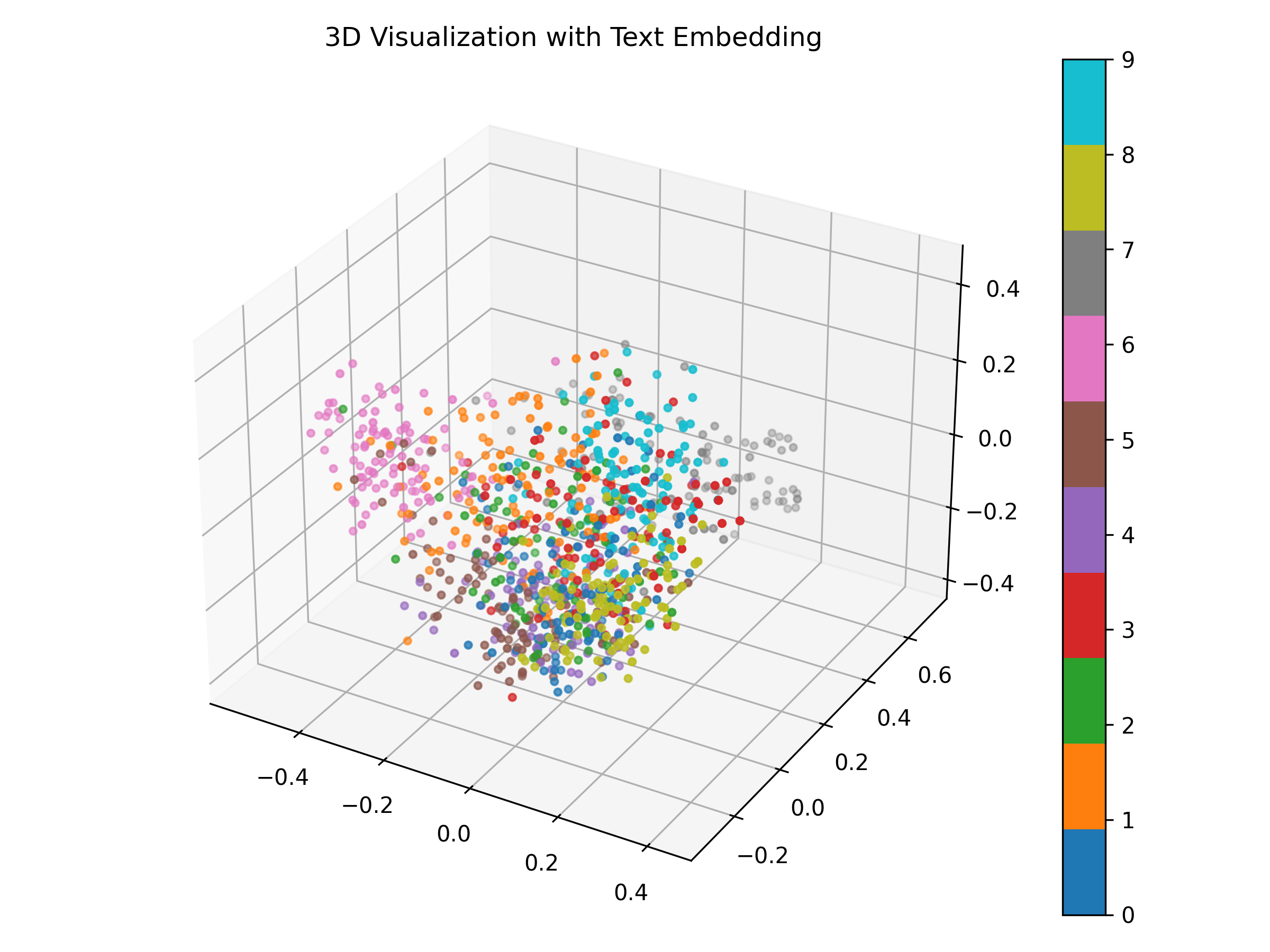}
    \caption{Clustering based on the sentence embedding. We visualize  100 samples in each cluster.}
    \label{fig:cluster_embedding}
\end{figure}

As noted in \cite{ostapenko2024towards}, LoRA similarity can contribute to positive transfer. Their approach clusters tasks based on LoRA similarity, requiring the \textit{task ID} for each example and the pre-training of a collection of LoRA adapters before clustering. In contrast, we argue that text-level clustering is more convenient while achieving competitive results. To validate this, we first examine the embedding representation of each example. As shown in Fig. \ref{fig:cluster_embedding}, we select 100 examples per cluster and visualize their embeddings. The embeddings reveal that similar texts are grouped into the same clusters, demonstrating the quality of our embedding representations.

\subsection{TC-LoRA for Parameter Compression}
\label{sec:memory_usage}
\textit{Task singular vectors} (TSV) \cite{gargiulo2025task} can leverage the low-rank structure of per-layer task matrices to effectively compress the LoRA adapters. Given a known task index $h$, they set the task components to zero, then the merged weight is :

\begin{equation}
    \hat{\Delta} = \sum_{i=1}^{N} \mathbb{I}_{[i = h]} \sum_{r=1}^R \left( \sum_{i=1}^N a_{ir} \right) \textbf{b}_r \otimes \textbf{c}_r
= \sum_{r=1}^R a_{hr} \mathbf{b}_r \otimes \mathbf{c}_r
\end{equation}

This formula yields a low-rank approximation of $\Delta_h$. Only the top-$R$ components of the task-specific matrix are retrained. Increasing $R$ improves approximation but reduces the compression.

\begin{table}[h]
    \centering
    \begin{tabular}{l|cc}
    \toprule
      Library   & Acc. & Memory Usage\\
      \midrule
      \textbf{LoRA Lib} & 67.2& 2304.0MB\\
      \textbf{TSV-merging} & 66.6 & 2.37MB \\
      \midrule
      \textbf{C-LoRA} & 68.9 & 75.0MB \\
      \textbf{TC-LoRA} & 68.8 & 1.88MB \\
      \bottomrule
    \end{tabular}
    \caption{Accuracy and memory usage of different library approaches. Here we use the $R=1$ for CP decomposition and $k=1$ for TSV.}
    \label{tab:memory_usage}
    
\end{table}
Tab. \ref{tab:memory_usage} presents a comparison of accuracy and memory usage across different approaches. The results demonstrate that TC-LoRA maintains high performance, nearly matching C-LoRA accuracy, while drastically reducing memory usage compared to both LoRA Lib and C-LoRA. This suggests that TC-LoRA effectively compresses parameters without significant loss in accuracy, making it a promising approach for resource-constrained environments.

\bibliography{aaai2026}


\end{document}